\documentclass{ntupaper} 
\usepackage[]{url}  
\usepackage{graphicx} 
\usepackage{natbib}  
\usepackage{caption} 
\usepackage{algorithm}
\usepackage{algorithmic}
\usepackage{xparse}

\usepackage{newfloat}
\usepackage{listings}
\DeclareCaptionStyle{ruled}{labelfont=normalfont,labelsep=colon,strut=off} 
\floatstyle{ruled}
\newfloat{listing}{tb}{lst}{}
\floatname{listing}{Listing}

\usepackage{booktabs}
\usepackage{multirow}
\usepackage{array}

\usepackage{adjustbox}

\newcommand{\method}{\texttt{ScrambleToolBench}}
\usepackage{amsmath}
\usepackage{cleveref}
\usepackage{amssymb}
\usepackage[table]{xcolor}

\newcommand{\memup}[2]{%
  #1\,{\scriptsize\textcolor{green!45!black}
  {(\ensuremath{\uparrow}#2)}}}
\newcommand{\memdown}[2]{%
  #1\,{\scriptsize\textcolor{red!65!black}
  {(\ensuremath{\downarrow}#2)}}}
\newcommand{\memsame}[1]{%
  #1\,{\scriptsize\textcolor{gray}{(\ensuremath{=})}}}

\newcommand{\gainv}[1]{\textcolor{green!45!black}{+#1}}
\newcommand{\lossv}[1]{\textcolor{red!65!black}{-#1}}

\title{\method{}: Agents Search Exhaustively Even When Their Own Map Points to the Next Step}

\addaffiliation{lab}{DeCLaRe Lab, Nanyang Technological University, Singapore}

\addaffiliation{astar}{Agency for Science, Technology, and Research (A*STAR), Singapore}

\author[lab,astar]{Vernon Toh}
\author[lab]{Navonil Majumder}
\author[astar]{Zhengyuan Liu}
\author[astar]{Nancy F. Chen}
\author[lab]{Soujanya Poria}             

\correspondingauthor{Soujanya Poria}{soujanya.poria@ntu.edu.sg}

\declareorg{declare-lab}
\code{ScrambleToolBench}

\exportmetadata[2026]{doe2026ntupaper}  

\abstract{
To operate robustly in open-world environments, autonomous agents should be able to infer the behavior of unfamiliar systems through interaction alone, even in the absence of documentation.
However, existing tool-use benchmarks expose semantic tool schemas in static environments, allowing agents to rely on prior knowledge rather than autonomous discovery.
To address this limitation, we introduce \method{}, an interactive terminal benchmark designed to isolate behavioral reasoning. By removing semantic cues and enforcing a continuous task curriculum, the benchmark requires agents to uncover hidden tool behaviors entirely through trial-and-error interaction. 
The benchmark further introduces dynamic challenges, including mapping drift, stochastic action failures, and temporal execution windows, to evaluate whether agents can revise and adapt their hypotheses as the environment changes.
Our evaluation of state-of-the-art language models reveals that successful initial discovery does not translate into robust adaptation.
When faced with structural changes such as mapping drift, agents fail to use deductive strategies such as cycle tracing, and instead exhibit belief inertia or fall back to exhaustive search.
Increasing test-time reasoning only amplifies this expensive brute-force search rather than enabling deductive recovery.
While equipping agents with persistent memory reduces compounding errors, they remain unable to efficiently infer structural changes, highlighting a gap in current agent reasoning.
}

\begin{document}

\maketitle

\teaser*{\includegraphics[width=0.99\linewidth]{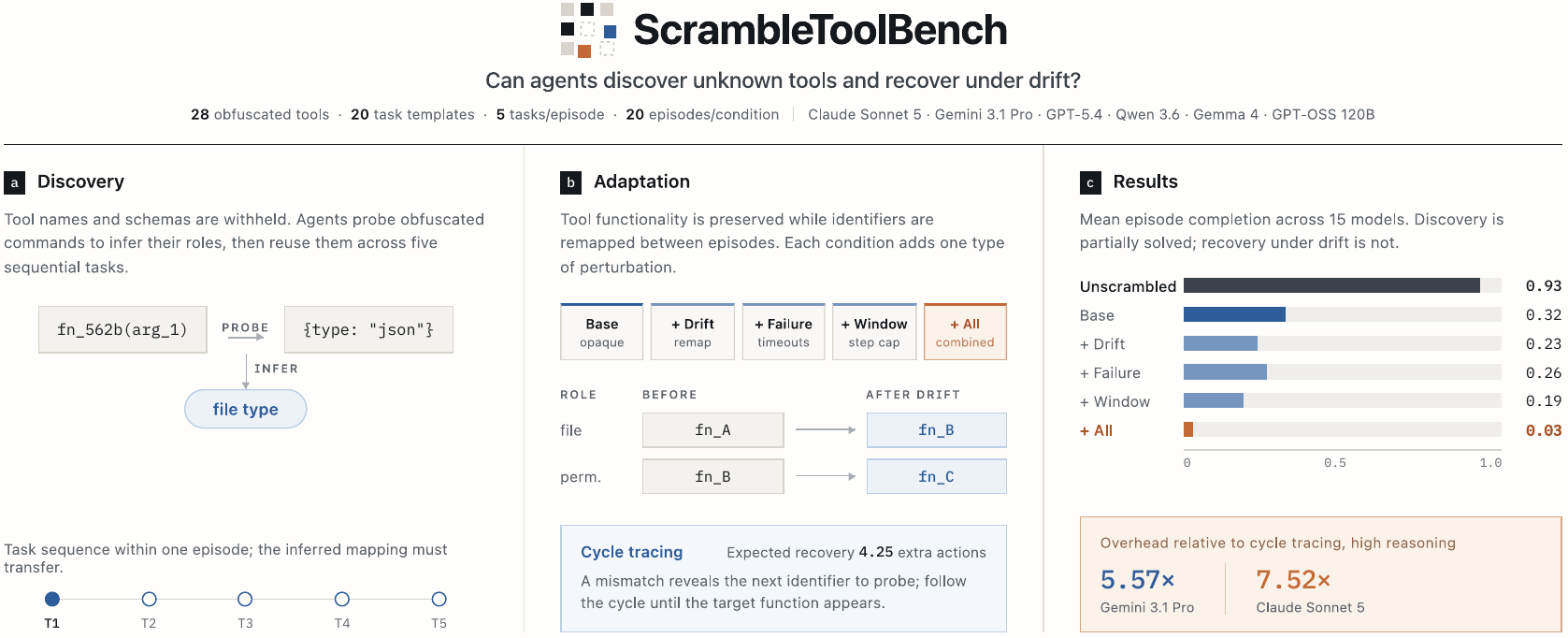}}

\section{Introduction}

Current large language model (LLM) agents demonstrate impressive capabilities when interacting with well-documented APIs and stable software environments \citep{wang2023voyageropenendedembodiedagent,cai2024largelanguagemodelstool}. 
However, this proficiency often relies heavily on semantic priors, i.e., understanding intuitive function names like \lstinline|read_file| or \lstinline|search|, rather than behavioral reasoning, which is the ability to deduce a tool's mechanics purely by observing its execution inputs and outputs. When these semantic cues are removed, or when the environment unexpectedly shifts, even state-of-the-art agents frequently struggle to adapt.
In many real-world systems, documentation may be incomplete, outdated, or unavailable.
Deployed APIs can diverge from their documented behavior, and transient failures such as network timeouts are common.
Agents must therefore infer hidden operational rules and required parameters through interaction, continuously updating their internal models as new evidence becomes available.

This requirement for autonomous discovery and adaptation exposes a significant gap in current evaluation frameworks. 
While recent interactive benchmarks have advanced the field by requiring agents to probe unfamiliar interfaces \citep{bandi2026mcpatlaslargescalebenchmarktooluse,luo2025mcpuniversebenchmarkinglargelanguage,hallinan2026opaquetoolsbenchlearningnuancestool} or navigate stateful environments \citep{jimenez2024swebench,xie2024osworldbenchmarkingmultimodalagents,merrill2026terminalbenchbenchmarkingagentshard,zheng2025lifelongagentbenchevaluatingllmagents,xi2026agentgym2benchmarkinglargelanguage}, they generally assume the underlying environment remains structurally stable once the tools are understood. 
Moreover, by retaining standard API naming conventions, these benchmarks effectively test an LLM's memorized knowledge rather than its capacity for deductive reasoning.
Consequently, they fail to evaluate an agent's ability to adaptively revise its established mental models when faced with dynamic environmental shifts, such as temporal execution windows, stochastic action failures, and underlying mapping drift. 
Reliable tool use requires agents to correctly identify the cause of unexpected failures, distinguishing transient noise from structural changes, and update their beliefs without relearning the entire environment from scratch.

\begin{figure*}[t]
    \centering
    \includegraphics[width=0.99\linewidth]{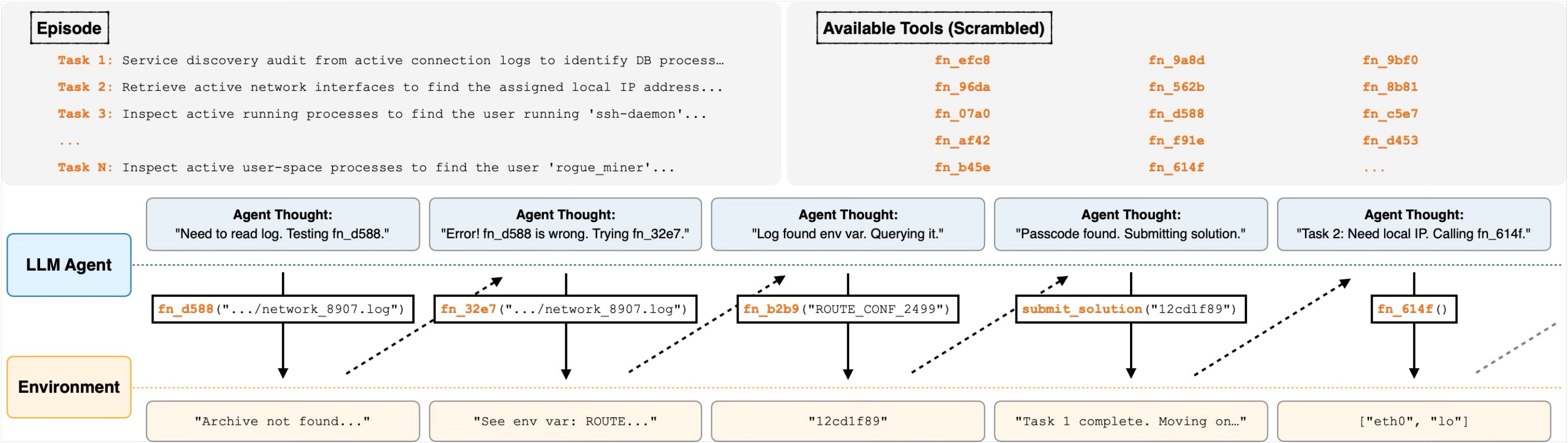}
    \caption{Overview of a \method{} evaluation episode. The agent is presented with a sequential curriculum of $N$ tasks (top-left) but has no prior semantic knowledge of the available tool action space (top-right), where all system tools are obfuscated with generic identifiers (such as \lstinline|fn_efc8|). To progress, the agent dynamically reasons, forms hypotheses, and probes commands in the stateful environment (bottom loop). For example, while attempting Task 1, the agent initially hypothesizes that \lstinline|fn_d588| reads files but receives an error. Recognizing the failure, it adapts and calls \lstinline|fn_32e7|, uncovering a clue pointing to an environment variable. Using this newfound context, the agent queries the variable via \lstinline|fn_b2b9|, successfully retrieving the required passcode. Upon submitting this passcode, the environment confirms Task 1 is complete. The agent then transitions into Task 2 in the same episode, immediately probing a new tool, \lstinline|fn_614f|, to discover network interfaces.}
    \label{fig:cryptex_overview}
\end{figure*}

To address this evaluation gap, we introduce \method{}, a novel interactive terminal benchmark designed to isolate deductive reasoning from pre-trained memorization (illustrated in Figure~\ref{fig:cryptex_overview}). 
We strip away all semantic priors by populating the environment with obfuscated command signatures (e.g., replacing \lstinline|read_file(path)| with randomized identifiers like \lstinline|fn_x9(arg_1)|). 
This acts as a worst-case scenario that forces agents to actively probe the environment to discover hidden operational behaviors across a continuous, progressive task curriculum. 
In contrast to previous benchmarks (Table~\ref{tab:benchmark_comparison}), \method{} explicitly tests the robustness of these learned hypotheses by introducing dynamic challenges such as stochastic action failures and mapping drift.
By evaluating agents under dynamic and uncertain conditions, the benchmark uncovers failure modes that remain hidden in static environments.

We evaluate state-of-the-art language models on \method{} to analyze their discovery capabilities and adaptability in non-stationary environments. 
Our empirical results demonstrate that while leading models achieve strong initial tool discovery in static environments, they struggle to adapt to mapping drift and transient failures. 
Under the combined effect of these challenges, the aggregate completion rate falls from 93\% to only 3\%.
Our analysis reveals that while models successfully build and maintain an internal map of tool behaviors, they fail to leverage this map when the environment shifts.
Rather than using deductive reasoning to trace permutation cycles, agents show signs of belief inertia or fall back on exhaustive search.
This demonstrates that agents struggle to leverage their previously established knowledge to deduce new rules and recover from unexpected changes.
Furthermore, we find that simply scaling test-time reasoning effort does not close this gap. While higher reasoning modes improve completion rates, they do so by enabling more exhaustive search at a significantly higher token cost, rather than by triggering deductive recovery strategies.
We further demonstrate that augmenting agents with a persistent memory decouples learning from execution, reducing compounding errors and representing an important step toward more resilient autonomous agents.

Our main contributions are as follows:
\begin{itemize}
    \item We introduce \method{}, a novel terminal-based benchmark for evaluating behavioral reasoning, the ability to form, test, and revise hypotheses about a tool's behavior through trial-and-error interaction rather than relying on semantic cues or prior knowledge. To isolate this capability, \method{} employs semantic obfuscation by anonymizing tool identifiers, parameters, and outputs.
    \item We propose a continuous task curriculum that evaluates hypothesis robustness by incorporating three dynamic environmental settings. First, mapping drift partially re-assigns tool identifiers between tasks. Second, stochastic action failures probabilistically return timeout errors for valid actions to simulate transient faults. Finally, temporal execution windows require specific action sequences to be completed within a strict step limit.
    \item Our evaluation reveals that leading language models face significant challenges in non-stationary environments where rules change unexpectedly. Under combined settings, aggregate completion rates drop from 93\% to 3\%, exposing a reasoning gap where agents, despite successfully mapping the initial environment, fail to logically deduce new mappings. Instead, they exhibit belief inertia, repeatedly testing invalidated hypotheses, or resort to exhaustive search rather than using recovery strategies such as cycle tracing. We show that increasing test-time reasoning effort fails to overcome this limitation, only resulting in more brute-force exploration.
    \item We demonstrate that externalizing state via a persistent memory provides a promising mechanism for resilience by preventing redundant re-discovery loops. However, we find that the effectiveness of persistent memory depends heavily on regular memory pruning, as stale beliefs can otherwise persist. Furthermore, despite using the external memory, agents fail to effectively track state to discover the cycle tracing strategy, incurring a significant action token overhead to solve tasks under drift.
\end{itemize}

\begin{table*}[t]
    \centering
    \resizebox{0.95\linewidth}{!}{
    \begin{tabular}{l c c c c c c}
        \toprule
        \multirow{2}{*}{\textbf{Benchmark}} & \textbf{Active Tool} & \textbf{Obfuscated Tool} & \textbf{Continuous} & \textbf{Progressive Task} & \textbf{Stochastic} & \textbf{Dynamic Tool} \\
        & \textbf{Discovery} & \textbf{Semantics} & \textbf{Interactive Session} & \textbf{Curriculum} & \textbf{Failures} & \textbf{Drift} \\
        \midrule
        ToolBench & $\times$ & $\times$ & $\times$ & $\times$ & $\times$ & $\times$ \\
        MCP-Atlas & \checkmark & $\times$ & \checkmark & $\times$ & $\times$ & $\times$ \\
        MCP-Universe & \checkmark & $\times$ & $\times$ & $\times$ & $\times$ & $\times$ \\
        OpaqueToolsBench & \checkmark & \checkmark & $\times$ & $\times$ & $\times$ & $\times$ \\
        SWE-bench & $\times$ & $\times$ & \checkmark & $\times$ & $\times$ & $\times$ \\
        OSWorld & $\times$ & $\times$ & \checkmark & $\times$ & $\times$ & $\times$ \\
        Terminal-Bench 2.0 & $\times$ & $\times$ & \checkmark & $\times$ & $\times$ & $\times$ \\
        LifelongAgentBench & $\times$ & $\times$ & \checkmark & \checkmark & $\times$ & $\times$ \\
        AgentGym2 & \checkmark & $\times$ & \checkmark & $\times$ & \checkmark & $\times$ \\
        \midrule
        \textbf{\method{} (Ours)} & \textbf{\checkmark} & \textbf{\checkmark} & \textbf{\checkmark} & \textbf{\checkmark} & \textbf{\checkmark} & \textbf{\checkmark} \\
        \bottomrule
    \end{tabular}
    }
    \caption{Comparison of \method{} with existing tool-use and agentic benchmarks across key evaluation dimensions.}
    \label{tab:benchmark_comparison}
\end{table*}

\section{The \method{} Benchmark}

An overview of an evaluation episode in \method{} is illustrated in \Cref{fig:cryptex_overview}. Rather than relying on plaintext descriptions of APIs and parameters, the agent must interactively discover tool behaviors by testing behavioral hypotheses across a sequential curriculum of stateful tasks.

\subsection{Problem Setting}

To capture the interactive discovery required in real-world systems, we formulate tool discovery as a sequential process within a stateful environment $\mathcal{E}$.
An agent interacts with the system by executing commands $(a, \mathbf{x})$, where $a$ is a command identifier and $\mathbf{x}$ maps argument keys to values.
Executing a command with valid keys $\mathcal{K}_a$ yields an observation $o$ and potentially transitions the state via $(s', o) = \text{Step}(s, a, \mathbf{x})$.
While target tools are state-preserving queries ($s' = s$), environment state transitions occur through task progression, solution submissions, or initiating temporal execution windows.

We divide the command space into plaintext meta-commands $\mathcal{M}$ and obfuscated target commands $\mathcal{C}$ to enforce empirical exploration.
Meta-commands $\mathcal{M} = \{\text{submit\_solution}, \text{skip\_task}, \text{end\_episode}\}$ provide known behaviors for managing the task curriculum.
In contrast, target commands $c \in \mathcal{C}$ conceal their schemas and operational behaviors. We emphasize that this semantic obfuscation is a worst-case scenario designed to isolate behavioral reasoning. 
In real-world systems, APIs often exhibit misleading, partial, or drifted semantics where over-reliance on semantic priors can lead to failures.
Invoking $c$ with invalid keys yields a schema mismatch error detailing the expected keys $\mathcal{K}_c$, forcing the agent to empirically deduce tool semantics through structured interaction.

We evaluate knowledge retention by structuring the evaluation as a continuous episode across a sequential curriculum of $N$ tasks $\mathcal{T} = \{T_1, \dots, T_N\}$.
For each task $T_i$, the agent receives a text-based goal $G_i$ (for example, ``\textit{Locate the corrupted log file and submit its ID}'') and a strict budget of $B$ actions. 
To successfully complete $T_i$, the agent must discover the required tool behaviors, retrieve the target information, and submit it using the \lstinline|submit_solution| meta-command. 
The environment evaluates the submission using exact string match against the actual ground-truth solution.
The budget is cumulative across attempts. If an agent uses the \lstinline|skip_task| meta-command and the task is re-queued, the actions used in prior attempts carry over.
If this cumulative budget $B$ is exhausted, the task is recorded as a failure and the evaluator progresses the agent to the next task in the curriculum.
By maintaining a continuous context history $\mathcal{H}_t$ across tasks, this design tests whether the agent can efficiently reuse discovered tool knowledge without exceeding its action budget.

\subsection{Obfuscation and Scrambling Mechanics}

To prevent models from bypassing discovery via semantic priors, \method{} enforces empirical exploration by anonymizing target command schemas.
At the beginning of each episode, we instantiate a bijective mapping $\Phi$ that obfuscates all target commands $c \in \mathcal{C}$ across four dimensions, while keeping meta-commands $\mathcal{M}$ in plaintext as control anchors.
This design ensures agents cannot infer behavior from linguistic cues and must dynamically map commands through execution: (i) \textbf{Function Identifiers} are mapped to randomized aliases (such as \lstinline|fn_3d8a|) and shuffled in the initial schema, which completely omits parameter definitions to force initial blind execution; (ii) \textbf{Parameter Keys} are assigned generic names (such as \lstinline|arg_0|) and shuffled per function, with their requirements and type hints (e.g., \lstinline|str|, \lstinline|int|) revealed only through plaintext schema mismatch errors; (iii) \textbf{Output Fields} are obfuscated with function-specific prefixes (such as \lstinline|out_c2b|) to hide predictable structural fields; and (iv) \textbf{Execution Status} is mapped to two randomized tokens per episode representing success and failure, with a success field systematically injected into all valid responses, forcing agents to empirically deduce execution outcomes.

\subsection{Task Taxonomy and Evaluation Protocol}
To prevent agents from memorizing fixed solution trajectories, \method{} separates \emph{what an agent can do} from \emph{what it is asked to do}. Specifically, the environment consists of a fixed set of API tools and a collection of procedural task templates. Evaluation episodes are generated by combining these two components, enabling many different tasks to be created from the same underlying toolset.

\textbf{(i) Core API Tools.} The environment provides $28$ core tools that support system operations, including file system management (e.g., \lstinline|list_dir|), network diagnostics (e.g., \lstinline|network_ping|), and data processing. It also includes three protected meta-commands (\lstinline|submit_solution|, \lstinline|skip_task|, and \lstinline|end_episode|) that control task execution and episode management. To ensure deterministic and efficient evaluation, $\mathcal{E}$ is implemented as a lightweight Python-based simulator instead of a live Docker-based environment.

\textbf{(ii) Procedural Task Templates.} On top of the shared API tools, \method{} defines $20$ procedural task templates that capture a diverse range of system administration, diagnostic, and cybersecurity scenarios.
When these templates are procedurally instantiated per episode, placeholders for targets (e.g., specific file paths, ports, or usernames) are dynamically populated, and the environment simulator computes the corresponding ground-truth solution for the generated state.

For each evaluation episode, the benchmark samples a sequential curriculum of $N$ tasks from this template pool. Since some API tools are reusable across different task templates, agents must both discover tool behaviors when they are first encountered and retain that knowledge for subsequent tasks. 
Therefore, this evaluates not only an agent’s ability to solve individual tasks, but also its ability to accumulate and reuse operational knowledge throughout an episode. A complete list of the available API tools and task templates is provided in \Cref{app:api_functions,app:task_taxonomy}.

\subsection{Environmental Dynamics and Difficulty Scaling}

To simulate the unpredictability of production systems, \method{} introduces three independent environmental dynamics that isolate specific failure modes in agent reasoning.

\paragraph{Mapping Drift.} To mimic software updates or API migrations, \method{} implements mapping drift between curriculum tasks. 
The scrambling mapping $\Phi$ undergoes a partial permutation parameterized by a drift fraction $\rho_{\text{drift}} \in [0, 1]$ (for example, $\rho_{\text{drift}} = 0.50$ re-assigns identifiers for half the active functions). 
To ensure the agent retains the ability to navigate the environment, meta-commands such as \lstinline|submit_solution|, \lstinline|skip_task|, and \lstinline|end_episode| are excluded from this process.
The permutation is implemented as a cyclic permutation among the selected functions, guaranteeing that the chosen tool always receives a new identifier.
Upon encountering unexpected failures, the agent should detect distribution shifts, invalidate stale mental models, and re-explore drifted mappings while preserving its learned dictionary.

\paragraph{Stochastic Action Failure.} To simulate execution faults like network timeouts, each valid execution has a failure probability $p_{\text{fail}} \in [0, 1]$, returning a timeout message instead of the expected output. 
As with mapping drift, meta-commands are excluded from these failures to preserve reliable environment navigation.
This tests hypothesis robustness by requiring the agent to attribute transient errors to environmental noise and retry rather than prematurely discarding a correct hypothesis.
These stochastic action failures act as environmental noise and do not consume the agent's action budget.

\paragraph{Temporal Execution Windows.} To evaluate execution efficiency and penalize brute-force exploration, specific tasks enforce a strict window of $k$ actions after a trigger command is invoked. The agent must complete the target sequence and submit its solution within this $k$-step limit. 
The environment makes this temporal limit observable by appending the remaining action countdown to every execution feedback. If the window expires before a successful submission, the environment resets the state and regenerates all intermediate variables.
This tests procedural mastery by ensuring the agent efficiently executes learned procedures without relying on memorized transient values.
When stochastic action failures are active, they are treated as uncontrollable delays and thus do not count toward the $k$-step limit.

\begin{table*}[t]
\centering

\begingroup
\setlength{\tabcolsep}{2.5pt}
\renewcommand{\arraystretch}{0.96}

\begin{adjustbox}{max width=0.99\textwidth,center}
\begin{tabular}{
    @{}l
    *{10}{c}
    >{\columncolor{red!3}}c
    >{\columncolor{red!3}}c
    @{}
}
\toprule

\textbf{Model}
& \multicolumn{2}{c}{\textbf{Unscrambled}}
& \multicolumn{2}{c}{\textbf{Base}}
& \multicolumn{2}{c}{\textbf{+ Drift}}
& \multicolumn{2}{c}{\textbf{+ Failure}}
& \multicolumn{2}{c}{\textbf{+ Window}}
& \multicolumn{2}{c}{\textbf{\textcolor{red!65!black}{+ All}}} \\

\cmidrule(lr){2-3}
\cmidrule(lr){4-5}
\cmidrule(lr){6-7}
\cmidrule(lr){8-9}
\cmidrule(lr){10-11}
\cmidrule(lr){12-13}

& $P_{\mathrm{ep}}$ & $T_{\mathrm{avg}}$
& $P_{\mathrm{ep}}$ & $T_{\mathrm{avg}}$
& $P_{\mathrm{ep}}$ & $T_{\mathrm{avg}}$
& $P_{\mathrm{ep}}$ & $T_{\mathrm{avg}}$
& $P_{\mathrm{ep}}$ & $T_{\mathrm{avg}}$
& $P_{\mathrm{ep}}$ & $T_{\mathrm{avg}}$ \\

\midrule
\multicolumn{13}{l}{\textit{Open-Source Models}} \\[-1pt]

Qwen 3.6 35B-A3B
& 0.90 & 4.90
& 0.05 & 1.20
& 0.00 & 0.60
& 0.00 & 0.60
& 0.00 & 0.30
& 0.00 & 0.10 \\

Gemma 4 26B-A4B
& 1.00 & 5.00
& 0.00 & 0.40
& 0.00 & 0.30
& 0.00 & 0.45
& 0.00 & 0.30
& 0.00 & 0.00 \\

Gemma 4 31B
& 1.00 & 5.00
& 0.00 & 1.50
& 0.00 & 1.20
& 0.05 & 1.45
& 0.00 & 0.75
& 0.00 & 0.35 \\

GPT-OSS 20B
& 0.60 & 4.50
& 0.00 & 0.00
& 0.00 & 0.00
& 0.00 & 0.00
& 0.00 & 0.00
& 0.00 & 0.00 \\

GPT-OSS 120B
& 0.85 & 4.65
& 0.00 & 0.05
& 0.00 & 0.05
& 0.00 & 0.00
& 0.00 & 0.00
& 0.00 & 0.00 \\

\addlinespace[1pt]
Qwen 3.6 27B
& 0.95 & 4.95
& 0.55 & 4.00
& 0.25 & 3.15
& 0.35 & 3.85
& 0.15 & 2.90
& 0.00 & 0.75 \\

\quad\textit{+ Memory}
& -- & --
& \memup{0.60}{0.05} & \memup{4.20}{0.20}
& \memup{0.40}{0.15} & \memup{3.95}{0.80}
& \memsame{0.35}      & \memup{3.90}{0.05}
& \memup{0.35}{0.20} & \memup{3.45}{0.55}
& \memsame{0.00}      & \memup{1.55}{0.80} \\

\midrule
\multicolumn{13}{l}{\textit{Proprietary Models}} \\[-1pt]

GPT-5.4 Mini
& 1.00 & 5.00
& 0.00 & 0.05
& 0.00 & 0.00
& 0.00 & 0.00
& 0.00 & 0.10
& 0.00 & 0.00 \\

GPT-5.4
& 1.00 & 5.00
& 0.10 & 2.05
& 0.00 & 0.80
& 0.05 & 2.05
& 0.05 & 1.25
& 0.00 & 0.15 \\

Gemini 2.5 Flash
& 0.85 & 4.65
& 0.05 & 2.05
& 0.00 & 1.25
& 0.00 & 1.40
& 0.00 & 1.45
& 0.00 & 0.20 \\

Gemini 2.5 Pro
& 0.95 & 4.75
& 0.45 & 3.80
& 0.00 & 1.50
& 0.10 & 2.25
& 0.05 & 2.80
& 0.00 & 0.55 \\

Gemini 3.1 Flash Lite
& 0.80 & 4.80
& 0.00 & 0.05
& 0.00 & 0.30
& 0.00 & 0.10
& 0.00 & 0.10
& 0.00 & 0.00 \\

Claude Sonnet 5
& 1.00 & 5.00
& 1.00 & 5.00
& 1.00 & 5.00
& 1.00 & 5.00
& 0.70 & 4.70
& 0.00 & 2.85 \\

\addlinespace[1pt]
Claude Sonnet 4.5
& 1.00 & 5.00
& 0.60 & 4.50
& 0.35 & 4.10
& 0.50 & 4.35
& 0.40 & 4.00
& 0.00 & 1.30 \\

\quad\textit{+ Memory}
& -- & --
& \memup{0.85}{0.25} & \memup{4.80}{0.30}
& \memup{0.60}{0.25} & \memup{4.20}{0.10}
& \memup{0.65}{0.15} & \memdown{4.00}{0.35}
& \memup{0.45}{0.05} & \memup{4.25}{0.25}
& \memsame{0.00}      & \memup{1.50}{0.20} \\

\addlinespace[1pt]
Gemini 3.1 Pro
& 1.00 & 5.00
& 1.00 & 5.00
& 0.90 & 4.90
& 1.00 & 5.00
& 0.80 & 4.80
& 0.20 & 2.35 \\

\quad\textit{+ Memory}
& -- & --
& \memsame{1.00}      & \memsame{5.00}
& \memup{1.00}{0.10} & \memup{5.00}{0.10}
& \memsame{1.00}      & \memsame{5.00}
& \memup{0.90}{0.10} & \memup{4.90}{0.10}
& \memup{0.50}{0.30} & \memup{3.60}{1.25} \\

\addlinespace[1pt]
Gemini 3.5 Flash
& 1.00 & 5.00
& 1.00 & 5.00
& 0.90 & 4.90
& 0.85 & 4.85
& 0.65 & 4.65
& 0.25 & 4.05 \\

\quad\textit{+ Memory}
& -- & --
& \memsame{1.00}      & \memsame{5.00}
& \memup{0.95}{0.05} & \memup{4.95}{0.05}
& \memup{0.95}{0.10} & \memsame{4.85}
& \memup{0.80}{0.15} & \memup{4.80}{0.15}
& \memup{0.30}{0.05} & \memup{4.15}{0.10} \\

\midrule
\multicolumn{13}{l}{\textit{Aggregate Effects}} \\[-1pt]

\rowcolor{red!4}
\textbf{Mean, no memory}
& 0.93 & 4.88
& 0.32 & 2.31
& 0.23 & 1.87
& 0.26 & 2.09
& 0.19 & 1.87
& 0.03 & 0.84 \\

\rowcolor{red!7}
\quad $\Delta$ from Unscrambled
& -- & --
& \lossv{0.61} & \lossv{2.57}
& \lossv{0.70} & \lossv{3.01}
& \lossv{0.67} & \lossv{2.79}
& \lossv{0.74} & \lossv{3.01}
& \lossv{0.90} & \lossv{4.04} \\

\rowcolor{green!5}
\textbf{Mean memory effect}
& -- & --
& \gainv{0.08} & \gainv{0.13}
& \gainv{0.14} & \gainv{0.26}
& \gainv{0.06} & \lossv{0.08}
& \gainv{0.13} & \gainv{0.26}
& \gainv{0.09} & \gainv{0.59} \\

\bottomrule
\end{tabular}
\end{adjustbox}
\endgroup

\caption{Main results on \method{}. We report episode completion
rate ($P_{\mathrm{ep}}$) and average tasks solved ($T_{\mathrm{avg}}$)
over 20 five-task episodes. Small arrows give the paired change from
memory. Aggregate rows average 15 no-memory models and four memory
pairs. Under \textit{+ All}, mean no-memory performance falls from
$0.93/4.88$ to $0.03/0.84$, while memory recovers $0.09/0.59$.}
\label{tab:main_results}
\end{table*}

\section{Experiments}

\subsection{Experimental Setup}

We evaluate \method{} on a diverse set of open-source and proprietary frontier models. The open-source models include the Qwen 3.6 series (27B and 35B-A3B), Gemma 4 series (26B-A4B and 31B), and GPT-OSS series (20B and 120B) \citep{qwen3.6-27b,qwen3.6-35b-a3b,gemmateam2026gemma4,openai2025gptoss120bgptoss20bmodel}. Our proprietary evaluation covers OpenAI's GPT-5.4 and GPT-5.4 Mini, Google's Gemini 2.5 (Flash and Pro), Gemini 3.1 (Flash Lite and Pro), Gemini 3.5 Flash, as well as Anthropic's Claude Sonnet 4.5 and Claude Sonnet 5.

We evaluate the selected models across a progressive sequence of environmental difficulties to isolate their tool discovery capabilities. First, to establish a baseline tool-use capability, each model is evaluated in an unscrambled control environment. Following this control, we evaluate the models in the foundational scrambled environment (Base). We then introduce environmental dynamics in isolation to analyze their individual impact on discovery: stochastic action failure ($p_{\text{fail}} = 0.15$), mapping drift ($\rho_{\text{drift}} = 0.25$), and temporal execution windows ($k = 10$). Finally, we evaluate the models under the \textit{All} condition, which combines all three dynamics to simulate highly noisy and unstable systems. To ensure reproducibility, evaluations are standardized across different runs using a fixed random seed. Due to inference budget constraints, we limit evaluations to 20 episodes of 5 tasks each, with a budget of 100 inference steps per task. We also evaluate the \textit{+ Memory} baseline on a representative subset of models.

\paragraph{Memory-Enhanced Baseline (+ Memory).} To assess how persistent memory can mitigate compounding errors from drift and stochasticity, we introduce a memory-enhanced agent architecture. This baseline decouples long-term, abstract knowledge from transient execution states by maintaining two structured databases that persist across task boundaries and environment resets: \textit{Task Recipes}, which stores parameterized sequences of logical steps, and \textit{Tool Knowledge}, which maps scrambled command identifiers to their inferred behaviors, parameter constraints, and confidence levels. At each environment step, the framework serializes both databases into a structured JSON payload and injects them into the system prompt. To modify these databases, the agent co-generates structured memory updates by including an optional \lstinline|"memory_update"| dictionary alongside its standard \lstinline|"action"| key within its single JSON response payload. The environment framework parses this object and applies the partial updates to the persistent databases before executing the primary action. We provide an example of the memory state JSON schema and the corresponding model output format in \Cref{app:memory_schema}.

\subsection{Evaluation Metrics}

We evaluate an agent's capability for tool discovery and knowledge reuse across sequential, dependent tasks ($T_1, T_2, \dots, T_N$) using two dimensions of success. 
The first metric is the \textbf{Episode Completion Rate} ($P_{\text{ep}}$). This measures the proportion of episodes where the agent solves all $N=5$ tasks. Because tasks are sequential and share a continuous context, any task failure halts progress. Consequently, $P_{\text{ep}}$ serves as a stringent metric of reliability.
The second metric is the \textbf{Average Tasks Solved} ($T_{\text{avg}}$) per episode. This calculates the mean number of completed tasks, ranging from $0$ to $N$. Unlike the binary $P_{\text{ep}}$, $T_{\text{avg}}$ provides a granular progress measure. It effectively distinguishes between agents that fail immediately and those that solve multiple tasks before stalling due to state drift, error propagation, or context saturation.

\subsection{Main Results}

\paragraph{Reliance on Semantic Priors.}
Most evaluated agents depend heavily on semantic cues rather than active empirical exploration to solve tasks. In the unscrambled control environment, these priors allow models to reliably complete sequential tasks ($P_{\text{ep}} \ge 0.60$, $T_{\text{avg}} \ge 4.5$). However, removing them in the scrambled \textit{Base} condition causes most models (such as Gemma 4 26B-A4B, GPT-5.4 Mini, Gemini 3.1 Flash Lite) to collapse to a $0.00$ completion rate, exposing a lack of generalizable discovery capabilities. Conversely, frontier models like Gemini 3.1 Pro, Gemini 3.5 Flash, and Claude Sonnet 5 maintain perfect $1.00$ completion rates, successfully inferring tool behaviors through observation and exploration.

\paragraph{Impact of Environmental Dynamics.}
Environmental noise and non-stationarity disrupt agent adaptability. Mapping drift (\textit{+ Drift}), which re-assigns $25\%$ of tool identifiers mid-episode, causes completion rate drops (for instance, Qwen 3.6 27B falls from $0.55$ to $0.25$, and Gemini 3.1 Pro from $1.00$ to $0.90$). Stochastic action failure (\textit{+ Failure}) challenges less robust models to distinguish between incorrect hypotheses and noise, dropping Qwen to $0.35$ while frontier models remain unaffected. Temporal execution windows (\textit{+ Window}) constrain iterative experimentation, dropping Claude Sonnet 5 to $0.70$ and Qwen to $0.15$. Combining all three dynamics (\textit{+ All}) causes widespread collapse: previously capable models like Claude Sonnet 5 drop to $0.00$, leaving only Gemini 3.1 Pro ($0.20$) and Gemini 3.5 Flash ($0.25$) capable of solving tasks without external memory assistance.

\paragraph{Persistent Memory.}
Equipping agents with a persistent memory improves their resilience across all scrambled conditions. By logging discovered tool mappings and task recipes, agents better adapt to mapping drift ($+0.14$ aggregate completion rate increase) and tight execution windows ($+0.13$). For instance, memory boosts Qwen 3.6 27B's completion rates for \textit{+ Drift} ($0.25$ to $0.40$) and \textit{+ Window} ($0.15$ to $0.35$). Crucially, memory mitigates compounding errors in the highly unstable \textit{+ All} condition, recovering an average $+0.09$ completion rate and $+0.59$ tasks solved across models. Gemini 3.1 Pro's \textit{+ All} completion rate increases from $0.20$ to $0.50$ (tasks solved rising from $2.35$ to $3.60$), proving that decoupling exploration from execution prevents redundant re-discovery loops.

\begin{table*}[t]
    \centering
    \small
    \resizebox{0.99\textwidth}{!}{%
    \begin{tabular}{l cccccc !{\vrule width 1.2pt} cc cc !{\vrule width 1.2pt} cc !{\vrule width 1.2pt} cc cc}
        \toprule
        & \multicolumn{6}{c!{\vrule width 1.2pt}}{\textbf{Action Efficiency ($A_{\text{avg}}$)}} & \multicolumn{4}{c!{\vrule width 1.2pt}}{\textbf{Retry Rates}} & \multicolumn{2}{c!{\vrule width 1.2pt}}{\textbf{Failure Modes}} & \multicolumn{4}{c}{\textbf{Stale Tool Calls}} \\
        \cmidrule(lr){2-7} \cmidrule(lr){8-11} \cmidrule(lr){12-13} \cmidrule(lr){14-17}
        & & & & & & & \multicolumn{2}{c}{\textbf{+ Failure}} & \multicolumn{2}{c!{\vrule width 1.2pt}}{\textbf{+ All}} & & & \multicolumn{2}{c}{\textbf{+ Drift}} & \multicolumn{2}{c}{\textbf{+ All}} \\
        \cmidrule(lr){8-9} \cmidrule(lr){10-11} \cmidrule(lr){14-15} \cmidrule(lr){16-17}
        \textbf{Model} & \textbf{Unscr.} & \textbf{Base} & \textbf{+ Drift} & \textbf{+ Fail.} & \textbf{+ Win.} & \textbf{+ All} & \textbf{Imm.} & \textbf{Pers.} & \textbf{Imm.} & \textbf{Pers.} & \textbf{Budget Exh.} & \textbf{Early Exit} & \textbf{Avg} & \textbf{Total} & \textbf{Avg} & \textbf{Total} \\
        \midrule
        \multicolumn{17}{c}{\textit{Open-Source Models}} \\
        \midrule
        Qwen 3.6 35B-A3B & 32.3 & 112.0 & -- & -- & -- & -- & 5.1\% & 24.3\% & 3.8\% & 21.5\% & 2.1\% & 84.8\% & 0.32 & 26 & 1.53 & 92 \\
        Gemma 4 26B-A4B & 25.7 & -- & -- & -- & -- & -- & 0.3\% & 22.3\% & 1.6\% & 29.7\% & 3.0\% & 97.0\% & 0.53 & 50 & 0.16 & 12 \\
        Gemma 4 31B & 25.0 & -- & -- & 66.0 & -- & -- & 19.1\% & 27.1\% & 17.6\% & 25.2\% & 0.0\% & 100.0\% & 0.69 & 22 & 1.00 & 33 \\
        GPT-OSS 20B & 38.4 & -- & -- & -- & -- & -- & 0.0\% & 21.4\% & 0.0\% & 10.0\% & 0.0\% & 100.0\% & 0.20 & 4 & 0.08 & 2 \\
        GPT-OSS 120B & 32.9 & -- & -- & -- & -- & -- & 1.8\% & 21.9\% & 2.8\% & 20.0\% & 3.0\% & 97.0\% & 0.78 & 21 & 0.93 & 14 \\
        Qwen 3.6 27B & 26.5 & 129.7 & 203.0 & 127.1 & 161.3 & -- & 17.3\% & 43.3\% & 26.5\% & 45.8\% & 53.1\% & 46.9\% & 3.67 & 165 & 3.37 & 202 \\
        \quad\textit{+ Memory} & -- & 117.9 & 180.1 & 148.1 & 141.9 & -- & 9.9\% & 37.5\% & 17.4\% & 39.0\% & 56.1\% & 43.9\% & 2.05 & 90 & 2.74 & 156 \\
        \midrule
        \multicolumn{17}{c}{\textit{Proprietary Models}} \\
        \midrule
        GPT-5.4 Mini & 25.2 & -- & -- & -- & -- & -- & 9.6\% & 22.0\% & 1.5\% & 7.6\% & 0.0\% & 100.0\% & 0.02 & 2 & 0.11 & 6 \\
        GPT-5.4 & 25.0 & 86.0 & -- & 101.0 & 97.0 & -- & 47.3\% & 58.6\% & 34.0\% & 43.6\% & 0.0\% & 100.0\% & 0.83 & 35 & 0.65 & 20 \\
        Gemini 2.5 Flash & 25.9 & 94.0 & -- & -- & -- & -- & 2.2\% & 5.5\% & 1.0\% & 6.3\% & 1.0\% & 97.0\% & 0.00 & 0 & 0.10 & 3 \\
        Gemini 2.5 Pro & 26.1 & 92.3 & -- & 82.5 & 88.0 & -- & 35.9\% & 47.8\% & 35.0\% & 49.8\% & 0.0\% & 100.0\% & 1.49 & 58 & 1.45 & 42 \\
        Gemini 3.1 Flash Lite & 26.8 & -- & -- & -- & -- & -- & 12.0\% & 19.6\% & 13.4\% & 16.9\% & 0.0\% & 100.0\% & 0.23 & 7 & 0.26 & 7 \\
        Claude Sonnet 5 & 25.6 & 115.8 & 193.6 & 114.3 & 139.5 & -- & 71.8\% & 78.7\% & 83.5\% & 88.8\% & 97.5\% & 2.5\% & 2.33 & 70 & 5.32 & 165 \\
        Claude Sonnet 4.5 & 27.6 & 106.6 & 158.9 & 128.0 & 118.8 & -- & 18.1\% & 32.8\% & 7.3\% & 23.8\% & 49.3\% & 50.7\% & 2.38 & 93 & 1.76 & 86 \\
        \quad\textit{+ Memory} & -- & 92.9 & 172.7 & 108.5 & 129.1 & -- & 10.3\% & 29.0\% & 12.1\% & 30.7\% & 17.1\% & 82.9\% & 2.76 & 113 & 2.54 & 99 \\
        Gemini 3.1 Pro & 25.1 & 80.8 & 152.4 & 98.0 & 91.4 & 178.8 & 38.9\% & 59.4\% & 44.2\% & 59.8\% & 62.5\% & 37.5\% & 2.75 & 77 & 1.62 & 52 \\
        \quad\textit{+ Memory} & -- & 84.8 & 145.3 & 91.3 & 93.1 & 189.0 & 36.1\% & 67.3\% & 59.0\% & 75.6\% & 60.0\% & 40.0\% & 2.77 & 86 & 2.23 & 98 \\
        Gemini 3.5 Flash & 25.2 & 94.5 & 162.7 & 94.2 & 113.5 & 216.4 & 12.9\% & 18.7\% & 19.0\% & 33.1\% & 100.0\% & 0.0\% & 0.93 & 28 & 1.26 & 34 \\
        \quad\textit{+ Memory} & -- & 91.8 & 159.5 & 100.4 & 105.2 & 175.2 & 28.7\% & 42.7\% & 14.6\% & 25.8\% & 100.0\% & 0.0\% & 0.44 & 12 & 0.36 & 12 \\
        \bottomrule
    \end{tabular}%
    }
    \caption{Combined behavioral metrics across all models. \textbf{Action Efficiency} ($A_{\text{avg}}$): Average total actions to complete all curriculum tasks, computed over successful episodes only; Unscr.\ denotes the unscrambled control setting; double-dashes (--) indicate zero successful episodes or no run. \textbf{Retry Rates}: Immediate and persistent retry rates under the stochastic action failure condition (\textit{+ Failure}, $p_{\text{fail}} = 0.15$) and the combined condition (\textit{+ All}). \textbf{Failure Modes}: Average distribution of Budget Exhaustion and Early Exit failure modes across all scrambled conditions. \textbf{Stale Tool Calls}: Absolute targeted stale calls under Drift alone and the combined (All) setting, tracking the number of times the model attempts to call an invalidated function name during a task.}
    \label{tab:combined_behavioral}
\end{table*}

\section{Analysis}

To better understand the cognitive processes, failure modes, and learning dynamics of the evaluated language model agents under environmental stress, we conduct a detailed analysis. 
Our analysis is structured around four key perspectives: quantifying adaptation, persistence, and belief inertia; assessing the ability of agents to discover efficient recovery strategies; evaluating the role of reasoning effort; and analyzing qualitative reasoning trajectories.

\subsection{Quantifying Adaptation, Persistence, and Belief Inertia}

To quantify these traits, our evaluation focuses on four primary dimensions of agent behavior which includes action overhead, retry persistence, failure mode distribution, and stale tool calling.

\subsubsection{Action Overhead}

Environmental scrambling inflates the action overhead required to solve tasks. In the unscrambled control setting, all models complete the curriculum efficiently in $\sim$25--27 actions, establishing a tight baseline where semantic cues enable direct tool invocation. However, obfuscating these cues in the scrambled \textit{Base} condition inflates costs by $3$--$5\times$. For example, Gemini 3.1 Pro's actions rise from $25.1 \to 80.8$, and Qwen 3.6 27B's actions rise from $26.5 \to 129.7$.

Adding environmental stressors further compounds this action overhead, with mapping drift causing the most severe increases. When invalidated mappings force agents to re-explore the environment (\textit{Drift}), actions rise by an additional $1.89\times$ for Gemini 3.1 Pro ($80.8 \to 152.4$).

Conversely, persistent memory acts as a buffer that reduces action overhead under stress by preventing redundant re-discovery loops. This efficiency gain is evident in Qwen 3.6 27B under the \textit{Window} condition, where actions decrease from $161.3 \to 141.9$. Similarly, Gemini 3.5 Flash under the combined \textit{All} condition sees a $19\%$ reduction in actions ($216.4 \to 175.2$).

\subsubsection{Stochastic Retry and Persistence}

Frontier models exhibit strong error attribution by maintaining high retry rates under stochastic action failure conditions ($p_{\text{fail}} = 0.15$). We measure this resilience through Immediate Retry Rate (re-execution immediately after a timeout) and Persistent Retry Rate (eventual re-execution later in the trajectory). Claude Sonnet 5 demonstrates this capacity by achieving 71.8\% immediate and 78.7\% persistent retry rates, correctly identifying timeouts as environmental noise rather than mapping failures. Similarly, GPT-5.4 and Gemini 3.1 Pro show robust resilience, with persistent retry rates of 58.6\% and 59.4\%, respectively.

In contrast, smaller open-source models struggle with error attribution and rapidly abandon hypotheses upon failure. For instance, GPT-OSS 20B exhibits a 0.0\% immediate retry rate. 
Failing to distinguish transient noise from actual errors, these models prematurely abandon correct hypotheses and trigger costly re-exploration.

Transitioning to the fully stressed \textit{+ All} condition degrades retry persistence across most models, as mapping drift makes it difficult for agents to determine whether a failure is due to transient noise or a structural change.
When tools both fail randomly and change mappings, models struggle to identify the cause of the failure. Consequently, GPT-5.4's persistent retry rate drops from 58.6\% to 43.6\%.

Providing agents with persistent memory leads to more stable retry behavior under combined stress. By externalizing schemas into memory, it helps agents recover from runtime failures while reinforcing correct tool mappings. For example, under the \textit{+ All} condition, enabling memory increases Gemini 3.1 Pro’s persistent retry rate from 59.8\% to 75.6\%.

\subsubsection{Failure Mode Distribution}

The distribution of failure modes reveals distinct behavioral profiles among models when they fail to complete the curriculum. We classify failures into two modes: Budget Exhaustion (depleting the maximum allowed steps) and Early Exit (the model calling \lstinline|end_episode()|). Highly sensitive models, such as GPT-5.4, GPT-5.4 Mini, and Gemini 2.5 Pro, fail via early exits (100\%), meaning they quickly give up when encountering schema mismatches. Conversely, persistent models like Gemini 3.5 Flash (100\% budget exhaustion) and Claude Sonnet 5 (97.5\%) persist in their attempts until they run out of steps.

Adding persistent memory alters these failure profiles, with the specific effect depending on the model's underlying tenacity. For models with mixed failure modes, like Claude Sonnet 4.5, memory shifts failures toward early exits, dropping its budget exhaustion rate from 49.3\% to 17.1\%. However, for models that already tend to exhaust their budget, like Gemini 3.1 Pro and Gemini 3.5 Flash, the failure mode remains stable at 60.0\% and 100\% budget exhaustion, respectively. This indicates that persistent agents use memory to sustain exhaustive search rather than to make earlier termination decisions.

\subsubsection{Stale Tool Calling}

We use Stale Tool Calls to measure a model's belief inertia when environmental rules change. A stale call occurs when an agent attempts to invoke a previously valid command that has been invalidated by mapping drift ($\rho_{\text{drift}} = 0.25$). Under the Drift condition, Qwen 3.6 27B exhibits the highest belief inertia with 3.67 stale calls per task (165 total). Claude Sonnet 4.5 and Claude Sonnet 5 follow closely at 2.38 and 2.33 calls per task, respectively. In contrast, Gemini 3.5 Flash demonstrates high flexibility, averaging only 0.93 stale calls per task.

The combined \textit{All} setting modifies this inertia dynamically based on the model's failure profile. For models like Claude Sonnet 5, the combined stressors cause stale calling to more than double, reaching 5.32 calls per task. Conversely, early-exiting models like GPT-5.4 and Gemini 3.1 Pro show a reduction in stale calls because they choose to terminate the episode rather than execute redundant commands.

Persistent memory generally mitigates belief inertia, although this benefit depends on the model's exploration policy. For instance, memory cuts Qwen 3.6 27B's stale calls from 3.67 down to 2.05 per task, and Gemini 3.5 Flash's from 0.93 to 0.44. However, memory can occasionally be detrimental: Claude Sonnet 4.5 experiences an increase in stale calls (from 2.38 to 2.76) when using memory, suggesting that persistent databases can sometimes lock in outdated mappings if the agent's update policy is flawed.

\subsection{Cycle Tracing: A Cheap Recovery Strategy for Mapping Drift}
\label{sec:ceiling}

Mapping drift greatly increases the number of actions used by the agents
(Table~\ref{tab:combined_behavioral}). 
We show that a moved function can be recovered by tracing a chain of identifiers, a strategy we term Cycle Tracing. In our benchmark, this requires at most six additional calls per drift event, and can be executed without the agent knowing the permutation size or that the drift is strictly a permutation.

\subsubsection{Setup}

Let $\mathcal{F}$ be the set of $M$ tools/functions and $\mathcal{N}$ be the set of corresponding $M$ obfuscated
identifiers. At task $t$, their mapping is
\[
\Phi_t:\mathcal{F}\rightarrow\mathcal{N}.
\]
Calling identifier $n$ executes function $\Phi_t^{-1}(n)$. Before the next task $t+1$,
a drift applies a permutation $\psi_t$:
\begin{equation}
\Phi_{t+1}=\psi_t\circ\Phi_t.
\label{eq:drift}
\end{equation}

Specifically, between two tasks, a drift event selects $k=7$ of the $M=28$ identifiers and
applies a new permutation to them. Thus, a five-task episode contains four
successive drift events, and the identifiers selected at different inter-task boundaries
may overlap. Each event forms one cycle: if its identifiers are
$n_0,\ldots,n_{k-1}$, then
\[
\psi_t(n_i)=n_{(i+1) \bmod k}.
\]
The other identifiers remain unchanged at that boundary.

Assume that the agent discovered the full mapping in Task~1 and keeps a table
in both directions. The output keys identify which function answered a call,
even after its identifier changes. We first consider one task boundary where
the stored table is current before drift, so $\hat\Phi=\Phi_t$.

\subsubsection{Cycle tracing recovery}
In this section, we present a deterministic algorithm to solve the problem. Suppose the agent calls identifier $n$ and observes function $f'$. 
It maintains a table that stores the bijective mapping between functions and their identifiers. 
Its table
shows that $f'$ previously had an identifier
\[
x'=\hat\Phi(f').
\]
If the table is current, then $f'$ moved from $x'$ to $n$. Therefore,
$n=\psi_t(x')$, or equivalently $x'=\psi_t^{-1}(n)$. As such, calling $x'$ will trace the
permutation one step backwards.

This next call is still useful when the agent's table is a patchwork of
observations from different tasks. If $n\neq x'$, bijectivity guarantees that
$f'$ cannot currently be at both identifiers. The stored entry for $x'$ is
therefore stale in the table. 
Naturally, $x'$ is a valid identifier to probe next. A current table is
needed only for the exact identity $x'=\psi_t^{-1}(n)$ and for the guaranteed cost as below.

Now, let $f$ be the function required by the task and $x=\hat\Phi(f)$ be its old
identifier. Let $\ell$ be the length of the cycle containing $x$. Starting at
$x$ and repeatedly applying the rule above traces that cycle backward and
reaches the new identifier of $f$ after $\ell$ calls. The last call executes
$f$ and replaces the call already needed to solve the task, so recovery uses
$\ell-1$ extra calls.

For a permuted function in our benchmark, the event forms one cycle over all $k$
selected identifiers. Consequently, $\ell=k=7$ and recovery uses $k-1=6$ extra calls. An
unchanged function has $\ell=1$ and needs no recovery. 
Note that the agent does not need
to know $\ell$ or $k$. And it stops when $f$ appears. The same calls reveal the new
mapping of the cycle, so the table can be updated immediately.

Therefore, no separate drift check is needed. The first call to $x$ is part of the task.
If it still executes $f$, the task continues normally. Otherwise, its response gives the first step of cycle tracing. 
A reasonable policy is to test the simplest local explanation first. When an identifier returns a different but previously discovered function, the agent can use its bijective map and Task~1 exploration traces to find that function’s previous identifier and probe it next. It can continue this process until it finds the function required by the task, while updating the map after every call. This does not require an additional call to identify the returned function, as the current call provides that evidence. Under this algorithm, for the 19 functions with arguments, Task~1 records both a validation-error fingerprint and a behavioral response. After drift, an argument mismatch naturally does not require the function to be discovered again. The stable output keys in the response can be matched against the earlier exploration trace to identify the function now occupying that identifier. Thus, the call that reveals the mismatch also identifies the next step of Cycle Tracing, and we do not charge an additional discovery call. If the function cannot be identified or its previous identifier is absent from the map, the agent can fall back to a more exhaustive search. Even if an agent does not discover this strategy immediately, it is reasonable to test whether it learns the repeated displacement pattern over later tasks. Our results show that the evaluated agents generally do not.

\subsubsection{Expected cost}

Each benchmark drift moves $k=7$ of the $M=28$ identifiers in one cycle. Each
task requires $r=4$ functions. If the moved identifiers are chosen uniformly,
the probability that at least one required function moved is
\[
1-\frac{\binom{M-k}{r}}{\binom{M}{r}}.
\]
The expected recovery cost for one fresh task boundary is therefore
\begin{equation}
\mathbb{E}[\Delta]
=
\left(
1-\frac{\binom{M-k}{r}}{\binom{M}{r}}
\right)(k-1).
\label{eq:delta}
\end{equation}
For our setting,
\[
\mathbb{E}[\Delta]
=
\left(
1-\frac{\binom{21}{4}}{\binom{28}{4}}
\right)6
=4.25
\]
extra actions~\footnote{Equation~\ref{eq:delta} assumes that the stored table is current immediately
before one drift event. If unresolved mappings remain from earlier tasks, the
relevant cycle may be longer and the cost can differ.
}.

A blind search would be more expensive here. 
After the old identifier gives the wrong
function, a random scan of the other identifiers uses $M/2$ extra calls on
average. The cycle tracing is cheaper for locating one function when
\[
k-1<\frac{M}{2}.
\]
For $M=28$ and $k=7$, this condition is satisfied. The cycle tracing remains valid when this inequality does not hold, but blind search may then be cheaper on average.

Thus, the cyclic structure of Drift provides a cheap way to recover the mapping. We could make Drift more difficult, which would likely reduce the performance of frontier models such as Claude Sonnet 5 and Gemini 3.1 Pro under a fixed action budget. However, such a result would be unsurprising: with a sufficiently large budget, agents could still recover the mapping through exhaustive search. Our goal is different. We ask whether agents can recognize and exploit the permutation structure using a simple method such as cycle tracing. Surprisingly, they do not.

\subsubsection{Observed cost}

Cycle tracing can guarantee recovery from Task~2 onward if the agent has
identified every function before the first drift. We define \emph{full
operational discovery} as obtaining a behavior-bearing response from every
environment function and pairing its response fingerprint with its current
identifier. The output keys provide the fingerprint because they remain
attached to the function when its identifier changes.

One call is sufficient for each of the nine functions without arguments. For
each of the other 19 functions, one call reveals the hidden argument schema and
a second, correctly shaped call produces a function-specific
response.\footnote{The second call may return an application-level error, such
as \emph{File not found} or \emph{Host unreachable}. We still count it as a
behavior-bearing discovery response because it reveals the function's stable
output fingerprint and behavior beyond a generic schema mismatch. Any
additional calls needed to find a successful input depend on the agent's
choices and are included in the observed cost, not in the discovery reference.}
Full operational discovery, therefore, has the reference cost
\[
A_{\mathrm{disc}}^{\mathrm{ref}}=9+2(19)=47.
\]
These calls can include the four function calls needed for Task~1. Including
the submission, the Task~1 reference is therefore $48$ actions.

For each later task, we
calculate
\[
\frac{A_t^{\text{Drift}}-5}
     {\mathbb{E}[\Delta]}.
\]
With this, we compare the model's actions beyond the five solving actions with the
expected recovery cost. We pair episodes completed in both \textit{Base} and
\textit{+ Drift}.

\begin{table*}[t]
\centering
\small
\begin{tabular*}{\textwidth}{@{\extracolsep{\fill}}lrrrrrr}
\toprule
Task & Base obs. & Drift obs. & Solve ref. & Compared cost & Reference & Ratio \\
\midrule
\multicolumn{7}{l}{\textit{Gemini 3.1 Pro, high reasoning} ($n=19$ paired episodes)} \\
1 & 41.63 & 40.47 & ---  & 40.47 & 48.00 & 0.84$\times$ \\
2 & 16.47 & 30.00 & 5.00 & 25.00 &  4.25 & 5.89$\times$ \\
3 &  8.21 & 33.58 & 5.00 & 28.58 &  4.25 & 6.73$\times$ \\
4 &  8.95 & 26.79 & 5.00 & 21.79 &  4.25 & 5.13$\times$ \\
5 &  5.74 & 24.26 & 5.00 & 19.26 &  4.25 & 4.54$\times$ \\
\cmidrule(lr){1-7}
2--5 avg. & 9.84 & 28.66 & 5.00 & 23.66 & 4.25 & \textbf{5.57$\times$} \\
\midrule
\multicolumn{7}{l}{\textit{Claude Sonnet 5, low reasoning} ($n=19$ paired episodes)} \\
1 & 44.63 & 50.63 & ---  & 50.63 & 48.00 & 1.05$\times$ \\
2 & 35.32 & 56.68 & 5.00 & 51.68 &  4.25 & 12.17$\times$ \\
3 & 19.26 & 42.26 & 5.00 & 37.26 &  4.25 &  8.78$\times$ \\
4 & 17.79 & 40.68 & 5.00 & 35.68 &  4.25 &  8.40$\times$ \\
5 &  7.84 & 32.16 & 5.00 & 27.16 &  4.25 &  6.40$\times$ \\
\cmidrule(lr){1-7}
2--5 avg. & 20.05 & 42.95 & 5.00 & 37.95 & 4.25 & \textbf{8.94$\times$} \\
\midrule
\multicolumn{7}{l}{\textit{Claude Sonnet 5, medium reasoning} ($n=20$ paired episodes)} \\
1 & 50.05 & 48.00 & ---  & 48.00 & 48.00 & 1.00$\times$ \\
2 & 27.45 & 32.30 & 5.00 & 27.30 &  4.25 & 6.43$\times$ \\
3 & 17.65 & 37.45 & 5.00 & 32.45 &  4.25 & 7.64$\times$ \\
4 & 13.25 & 36.65 & 5.00 & 31.65 &  4.25 & 7.45$\times$ \\
5 &  9.95 & 38.90 & 5.00 & 33.90 &  4.25 & 7.98$\times$ \\
\cmidrule(lr){1-7}
2--5 avg. & 17.08 & 36.33 & 5.00 & 31.33 & 4.25 & \textbf{7.38$\times$} \\
\midrule
\multicolumn{7}{l}{\textit{Claude Sonnet 5, high reasoning} ($n=20$ paired episodes)} \\
1 & 50.35 & 45.85 & ---  & 45.85 & 48.00 & 0.96$\times$ \\
2 & 25.85 & 33.35 & 5.00 & 28.35 &  4.25 & 6.68$\times$ \\
3 & 20.00 & 38.30 & 5.00 & 33.30 &  4.25 & 7.84$\times$ \\
4 & 12.65 & 32.85 & 5.00 & 27.85 &  4.25 & 6.56$\times$ \\
5 &  6.95 & 43.20 & 5.00 & 38.20 &  4.25 & 9.00$\times$ \\
\cmidrule(lr){1-7}
2--5 avg. & 16.36 & 36.93 & 5.00 & 31.93 & 4.25 & \textbf{7.52$\times$} \\
\bottomrule
\end{tabular*}
\caption{Mean action cost including budget-failed attempts. Episodes with an
early task exit are excluded. For Task~1, Compared cost is the observed total,
Reference is the 48-action full-discovery reference, and Ratio compares the two.
A Task~1 ratio below one is not evidence of a better discovery strategy because
the models did not complete the mapping. For
Tasks~2--5, Compared cost is $A_t^{\mathrm{Drift}}-5$, Reference is
$\mathbb{E}[\Delta]$, and Ratio compares the observed recovery overhead with
the unrounded value of Eq.~\ref{eq:delta}.}
\label{tab:drift_cost}
\end{table*}

Both models solve all 20 Task~1 episodes at high reasoning, but they do not
finish discovery. By the end of Task~1, Gemini has successfully invoked
$20.75$ of the 28 environment functions on average and Sonnet has invoked
$21.90$. Neither model obtains a response from all 28 functions in any of these
drift runs. They continue discovering functions in later tasks, after their
stored observations have already been affected by drift.

Gemini 3.1 Pro and Claude Sonnet 5 achieve strong task success in both
\textit{Base} and \textit{+ Drift}, but neither model consistently finds the
cheaper cycle tracing strategy. This suggests that finding a cheaper search path may
depend on reasoning. To test this hypothesis, we compare Sonnet at low, medium,
and high reasoning in \Cref{tab:drift_cost}. At low reasoning, seven of the
$19$ retained drift episodes contain at least one task that reaches the
100-action limit. This falls to one of $20$ episodes at medium reasoning and
zero of $20$ at high reasoning. After including these failed attempts, Sonnet's
mean ratio is $8.94\times$ at low reasoning, compared with $7.38\times$ and
$7.52\times$ at medium and high reasoning.

We next test whether this improvement comes from following the chain. After
observing that an identifier changed, we check whether the model calls the next
identifier in the chain within three actions.

\begin{table}[t]
\centering
\small
\resizebox{0.7\columnwidth}{!}{%
\begin{tabular}{llcccc}
\toprule
Model & Think & Opportunities & Follows & Random & $p$ \\
\midrule
Gemini 3.1 Pro  & high   & 484 & 11.0\% & 10.8\% & 0.934 \\
\quad + Memory & high & 444 & 12.8\% & 10.9\% & 0.195 \\
Claude Sonnet 5 & low    & 564 & 14.0\% & 10.6\% & 0.009 \\
                 & medium & 590 & 11.9\% & 10.6\% & 0.328 \\
                 & high   & 604 & 14.1\% & 10.6\% & 0.006 \\
\bottomrule
\end{tabular}
}
\caption{Use of the recovery chain under drift, with the same early-exit exclusion. Follows is the fraction of opportunities in which the model calls the next identifier within three actions. Random is uniform selection among identifiers already mapped by the model.}
\label{tab:chain_follow}
\end{table}

Results in \Cref{tab:chain_follow} show that Gemini is no better than random selection. Sonnet's follow rate is $14.0\%$,
$11.9\%$, and $14.1\%$ at low, medium, and high reasoning. 
Increased reasoning therefore reduces failed searches, but it does not make Sonnet follow the cheaper chain more consistently.

\paragraph{Takeaways.} Recovering from drift requires two capabilities: maintaining the learned mapping and finding an efficient strategy for using it. Our memory experiment supports the first by giving the agent persistent, model-updated tables of identifier and function hypotheses and previously learned task procedures. Under \textit{Drift}, memory reduces Gemini 3.1 Pro's total actions from \(154.9\) to \(139.9\), exact repeated calls from \(38.2\) to \(24.6\), and later-task calls to identifiers outside the current seven-identifier drift set from \(76.6\) to \(66.3\). The number of first encounters with changed mappings remains similar (\(20.4\) versus \(19.3\)), as expected, because the agent must first call a changed identifier to detect the drift. The important decision factor comes after this observation. When the required entries are present, the memory table can be searched in reverse to find the previous identifier of the function returned by the changed call. However, memory does not lead the agent to make this deduction as Gemini follows the correct next identifier within three actions in only \(12.8\%\) of opportunities, compared with a \(10.9\%\) random baseline (\(p=0.195\)). The task-position results show the same difference between remembering a mapping and recovering it after drift. Under \textit{Base}, action cost decreases across Tasks~2--5 for both models at every reasoning level, which is consistent with the agents reusing mappings learned in earlier tasks. At high reasoning, Gemini retains this trend under \textit{Drift}, with a slope of \(-2.40\). Sonnet reverses it at medium and high reasoning, with slopes of \(+1.90\) and \(+2.41\), meaning that its cost increases as the episode proceeds. Its low-reasoning slope of \(-7.51\) should not be interpreted as successful transfer as ten task attempts reach the 100-action budget limit. Moreover, nine of them occur in the first three task positions, which inflates the early costs. For Sonnet, additional reasoning does not improve persistence, because it never quits, and its low-reasoning failures already exhaust the available budget. Instead, it reduces repeated searches. Tool coverage remains almost unchanged (\(27.9\), \(27.8\), and \(27.8\)), while calls to previously explored identifiers fall from \(164.2\) at low reasoning to \(133.9\) and \(131.7\) at medium and high reasoning. In Tasks~2--5, calls to identifiers outside the current drift set also fall from \(107.7\) to \(90.5\) and \(91.2\). This reduces budget-failed episodes from eight to one with medium and then zero with high, but does not bring Sonnet close to Cycle Tracing: its recovery cost remains \(7.38\times\) and \(7.52\times\) the reference at medium and high reasoning, and its chain-following rate does not improve with reasoning effort. Thus, memory improves recall, and additional reasoning makes search less wasteful, but neither reliably leads the agents to derive and use the cheaper recovery strategy.

\subsection{The Role of Reasoning Effort}

\begin{figure*}[t]
\centering

\begin{minipage}[t]{0.485\textwidth}
    \centering
    \includegraphics[width=\linewidth]{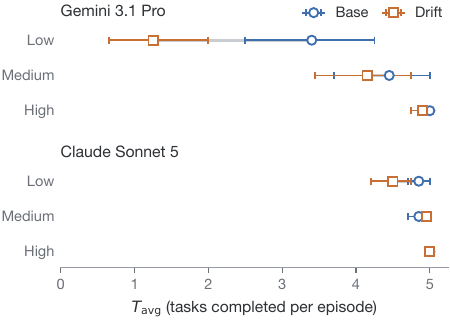}
    \vspace{-0.6em}

    {\footnotesize (a) Mean tasks completed, $T_{\mathrm{avg}}$.\par}
\end{minipage}
\hfill
\begin{minipage}[t]{0.485\textwidth}
    \centering
    \includegraphics[width=\linewidth]{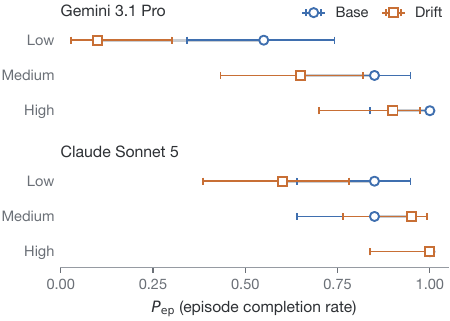}
    \vspace{-0.6em}

    {\footnotesize (b) Episode completion rate, $P_{\mathrm{ep}}$.\par}
\end{minipage}

\vspace{0.5em}

\begin{minipage}[t]{0.485\textwidth}
    \centering
    \includegraphics[width=\linewidth]{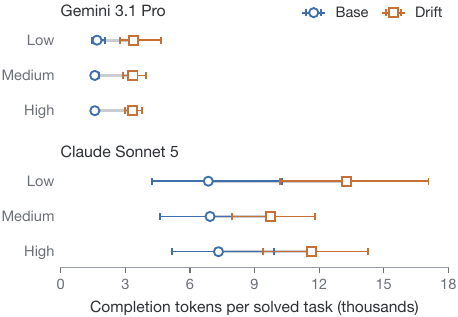}
    \vspace{-0.6em}

    {\footnotesize (c) Completion tokens per solved task.\par}
\end{minipage}
\hfill
\begin{minipage}[t]{0.485\textwidth}
    \centering
    \includegraphics[width=\linewidth]{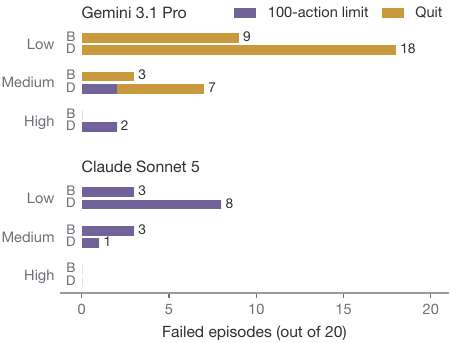}
    \vspace{-0.6em}

    {\footnotesize (d) Failed episodes by cause
    ($\mathrm{B}=\text{Base}$, $\mathrm{D}=\text{Drift}$).\par}
\end{minipage}

\caption{Reasoning-budget ablation over 20 five-task episodes per cell.
Panels (a)--(c) compare the scrambled baseline with mapping drift. Horizontal
intervals are 95\% bootstrap intervals for $T_{\mathrm{avg}}$ and token cost,
and Wilson intervals for $P_{\mathrm{ep}}$. Token cost includes completion
tokens from both successful and failed episodes and divides them by the total
number of solved tasks. Panel (d) separates episodes that reached the
100-action limit from episodes in which the agent quit.}
\label{fig:reasoning_ablation}
\end{figure*}

We report the results of Gemini 3.1 Pro and Claude Sonnet 5 at different
levels of reasoning effort in \Cref{fig:reasoning_ablation}. Higher reasoning
effort improves task completion for both agents. For Gemini 3.1 Pro,
$T_{\mathrm{avg}}$ increases from $3.40$ to $5.00$ in \textit{Base} and from
$1.25$ to $4.90$ under \textit{+ Drift}. Its episode completion rate
$P_{\mathrm{ep}}$ increases from $0.55$ to $1.00$ in \textit{Base} and from
$0.10$ to $0.90$ under drift. Claude Sonnet 5 shows a smaller improvement:
$T_{\mathrm{avg}}$ increases from $4.85$ to $5.00$ in \textit{Base} and from
$4.50$ to $5.00$ under drift, while $P_{\mathrm{ep}}$ increases from $0.85$
to $1.00$ and from $0.60$ to $1.00$, respectively. This is consistent with
the previous section: more reasoning reduces failed searches. However, it
does not show that either agent finds the cheaper recovery chain.

We also report completion tokens per solved task:
\[
C_{\mathrm{tok}}
=
\frac{\text{completion tokens from all episodes}}
     {\text{total solved tasks}}.
\]
The numerator includes tokens spent in failed episodes. At high reasoning,
the two agents achieve similar performance, but Claude Sonnet 5 uses
substantially more tokens. In \textit{Base}, it uses $7{,}326$ completion
tokens per solved task, compared with $1{,}584$ for Gemini 3.1 Pro
($4.6\times$). Under drift, it uses $11{,}652$ tokens, compared with
$3{,}332$ for Gemini ($3.5\times$).

Lower reasoning does not reliably reduce token cost. Gemini's cost remains
nearly unchanged across reasoning levels: $1{,}580$--$1{,}686$ tokens in
\textit{Base} and $3{,}332$--$3{,}382$ under drift. For Claude Sonnet 5
under drift, low reasoning is the most expensive setting, using $13{,}274$
tokens per solved task, compared with $9{,}741$ at medium reasoning and
$11{,}652$ at high reasoning.

As shown in \Cref{tab:drift_cost}, Claude Sonnet 5 also uses more actions
after Task~1. At high reasoning, its mean action count over Tasks~2--5 is
$16.36$ in \textit{Base} and $36.93$ under drift, compared with $9.84$ and
$28.66$ for Gemini 3.1 Pro. The agents also fail differently at low
reasoning. Gemini quits early in all of its failed episodes i.e., 9 in \textit{Base} and 18 under drift, with none reaching the 100-action limit.
Claude Sonnet 5 shows the opposite behavior in that all three of its \textit{Base}
failures and all eight of its drift failures reach the action limit, and it never quits early.

\subsection{Qualitative Insights into Agent Reasoning}

To understand the reasoning behind the numerical trends, we analyze the step-by-step actions of the agents. We identify three common patterns of behavior that explain how agents succeed or fail when facing uncertain environments.

\subsubsection{Repeated Errors and Stuck Beliefs}

Agents often fail because they struggle to update their internal plans after the environment changes. For example, Qwen 3.6 27B (Episode 6, Task 3) initially uses \lstinline|fn_fc40| to search files successfully. However, when mapping drift invalidates this tool, subsequent calls to \lstinline|fn_fc40| begin returning empty results (\lstinline|{"out_bc5": [], "out_3c0": "val_647"}|).

Rather than abandoning the invalidated tool, the agent gets stuck in a repetitive 7-step execution loop. In Step 3, the agent reasons: ``Let me try searching for archive files with different patterns.'' and calls \lstinline|fn_fc40|. In Steps 4 through 7, the agent repeatedly generates similar rationalizations (such as ``maybe the archive file has a numeric suffix...'') and continues calling the exact same invalidated function.

This repetitive loop exposes a critical weakness in how smaller models handle unexpected observations. Instead of updating its understanding of the available commands upon receiving empty results, the model maintains a stuck belief and merely tries the same action with minor variations.

\subsubsection{Using Memory to Stay on Track}

Models equipped with memory avoid redundant re-discovery loops by saving validated tool mappings to a persistent memory. For instance, in Episode 6, Task 6, Gemini 3.5 Flash begins the task with a previously saved memory injected into its system prompt:
\begin{quote}
\lstinline[basicstyle=\footnotesize\ttfamily\itshape]|{"fn_ecfe": "list_cron_jobs()", "fn_40cd": "Query user groups"}|
\end{quote}
This externalized knowledge allows the agent to immediately map semantic goals to correct actions. When given the instruction to ``Query the active automated crontab tasks'', the agent reasons:
\begin{quote}
\textit{``Active automated crontab tasks? We know fn\_ecfe does that (we recorded it in Task 1 memory). Let's call fn\_ecfe.''}
\end{quote}
Consequently, the agent successfully executes the correct tool on its very first turn.
Beyond mere storage, frontier models demonstrate the ability to actively manage and prune their memory to prevent confusion during environmental shifts. In Episode 6, Task 3, Gemini 3.1 Pro detects that a new mapping drift has occurred and decides to clear its memory:
\begin{quote}
\textit{``Wait, first, we need to know the obfuscated tool mappings for Task 3 because they might have been re-shuffled again! Let's clear the previous tool mapping except the task progress/critical details.''}
\end{quote}
This targeted pruning highlights how careful memory management helps agents separate high-level planning from low-level environmental details. By actively removing stale mappings, the agent avoids the belief inertia observed in models without effective memory policies.

\subsubsection{Handling Temporal Execution Windows and Recovering from Resets}

Temporal execution windows enforce strict limits ($k=10$ steps) that suppress exploration and demand precise action sequences once a trigger command is executed. If an agent fails to finish the required sequence within $k$ steps, the environment resets, discarding all transient progress. By observing agent trajectories, we identify two primary strategies for managing these temporal constraints: focusing on key actions and dynamically recovering from resets.

When faced with an active temporal execution window, successful agents immediately shift their behavior from exploration to targeted execution. They abandon diagnostic tools and focus solely on steps they know will succeed. For example, when Gemini 3.5 Flash triggers a temporal execution window in Episode 14, Task 2, it acknowledges the constraint:
\begin{quote}
\textit{``A temporal window of k=10 is now active. I must avoid all diagnostic or exploratory calls like list\_dir or get\_system\_info and immediately execute the 3-step sequence: decode the payload, verify the host IP, and submit the solution. Let's call fn\_dec2 first.''}
\end{quote}
As a result of this focused execution, the model successfully completes the sequence in just three steps, safely within the constraint.

\paragraph{Recovering from Resets and Confusion.}

Environmental resets create a clear divide in agent capabilities, differentiating models that monitor environment state from those that operate blindly. While weaker models fail to notice resets and attempt to reuse invalidated session variables, frontier models detect the state change and dynamically adjust their plans.

This dynamic recovery is evident when models choose to backtrack after a failure. In Gemini 3.1 Pro during Episode 9, Task 4, the agent encounters an invalid state error caused by an expired window and correctly diagnoses the issue:
\begin{quote}
\textit{``Wait, the environment returned an invalid state error. The active temporal window must have expired, resetting the server state and clearing my session variables. I cannot proceed with the final solution submission. I must backtrack to step 1, re-invoke the trigger command fn\_abc1 to open a new active window, and quickly re-execute the sequence. Calling fn\_abc1.''}
\end{quote}
This deliberate backtracking demonstrates that the agent continuously monitors its state under time pressure, allowing it to recover gracefully and ultimately finish the task.

\section{Related Work}

\paragraph{Tool-Use Benchmarks.}
Tool-use benchmarks evaluate the capacity of autonomous agents to interface with external systems and APIs.
While foundational evaluations assume static environments with explicit plaintext schemas \citep{xu2023toolmanipulationcapabilityopensource}, recent approaches test interactive discovery by forcing agents to probe unfamiliar interfaces \citep{bandi2026mcpatlaslargescalebenchmarktooluse,luo2025mcpuniversebenchmarkinglargelanguage} or infer the semantics of intentionally obfuscated tools \citep{hallinan2026opaquetoolsbenchlearningnuancestool}.
However, these frameworks rely on structural stability: they evaluate how well an agent can initially reverse-engineer a tool, but assume the tool's behavior remains constant once understood.
In contrast, \method{} disrupts this stability by injecting mapping drift and stochastic action failures into the obfuscated tool space, testing whether agents can rapidly adapt their internal abstractions when the environment itself unexpectedly changes.

\paragraph{Interactive Benchmarks.}
Interactive agent benchmarks assess long-horizon reasoning and continuous execution within stateful computing environments.
These frameworks test an agent's ability to navigate complex software workflows \citep{jimenez2024swebench}, operate real computer interfaces \citep{xie2024osworldbenchmarkingmultimodalagents}, or manage multi-step error recovery across extended terminal sessions \citep{merrill2026terminalbenchbenchmarkingagentshard,zheng2025lifelongagentbenchevaluatingllmagents,xi2026agentgym2benchmarkinglargelanguage}.
Although these benchmarks rigorously evaluate multi-step reasoning in noisy settings, they typically provide rich semantic feedback and maintain stable underlying mechanics, rarely requiring the agent to question its fundamental understanding of the environment.
Building on these stateful dynamics, \method{} strips away all semantic priors and introduces structural volatility, testing whether an agent can systematically revise its established mental models in response to dynamic environmental shifts.

\section{Conclusion}

This paper addresses the fragility of language model agents in unstable environments by proposing \method{}, a benchmark that isolates behavioral reasoning by removing semantic priors and introducing dynamic stress. The key idea is to evaluate autonomous tool discovery through scrambled mappings, stochastic action failures, and temporal execution windows, which exposes the over-reliance of current models on semantic cues. Experiments show that while most agents fail under combined stress, providing frontier models with persistent memory recovers task completion rates and reduces redundant loops. Our analysis reveals a reasoning gap where agents faced with mapping drift exhibit belief inertia and fail to recognize structural permutations, opting instead for exhaustive search over deductive strategies like cycle tracing. Furthermore, we demonstrate that scaling test-time reasoning effort fails to induce these deductive capabilities, instead enabling brute-force exploration at a higher token cost.

\bibliographystyle{plainnat}
\bibliography{discovery}


\appendix

\section{Memory-Enhanced Baseline Schema}
\label{app:memory_schema}
To evaluate whether externalizing learned representations improves long-horizon task efficiency, the memory-enhanced baseline (+ Memory) equips the agent with two structured databases: \textit{Task Recipes} and \textit{Tool Knowledge}.

At each environment step, the framework serializes both databases into a structured JSON payload and injects them into the system prompt. To modify these databases, the agent co-generates structured memory updates by including an optional \lstinline|"memory_update"| dictionary alongside its standard \lstinline|"action"| key within its single JSON response payload. 

An example of the JSON state provided to the model in the prompt is as follows:
\begin{lstlisting}
{
  "system_state": { ... },
  "memory": {
    "Task_Recipes": {
      "Task_1": "1. Call fn_a1 to get X. 2. Call fn_b2 with X."
    },
    "Tool_Knowledge": {
      "fn_a1": {
        "inferred_behavior": "Reads file from path",
        "parameters": {"arg_0": "file_path (str)"},
        "confidence": "high"
      }
    }
  }
}
\end{lstlisting}

The agent can update this memory by outputting the following JSON structure:
\begin{lstlisting}
{
  "action": {
    "command": "fn_a1",
    "arguments": {"arg_0": "/etc/hosts"}
  },
  "memory_update": {
    "Tool_Knowledge": {
      "fn_a1": {
        "inferred_behavior": "Reads file from path",
        "parameters": {"arg_0": "file_path (str)"},
        "confidence": "high"
      }
    }
  }
}
\end{lstlisting}

\section{Core API Functions}
\label{app:api_functions}

The API functions represent standard operations required across different task taxonomies:
\begin{itemize}
    \item \lstinline|list_dir(path: str)|: Lists all files and directories in a given virtual filesystem path.
    \item \lstinline|read_file(file_path: str)|: Reads the full text content of a file from the virtual filesystem.
    \item \lstinline|search_text(query: str)|: Globally searches across all files in the virtual filesystem for a plain-text substring.
    \item \lstinline|decode_base64(text: str)|: In-memory utility that decodes a Base64 string back into UTF-8 text.
    \item \lstinline|submit_solution(solution: str)|: Evaluates whether the proposed solution string matches the correct task solution.
    \item \lstinline|get_system_info(category: str)|: Retrieves high-level operating system and platform categories (e.g., 'os', 'hardware', 'env').
    \item \lstinline|network_ping(host: str)|: Measures network-layer reachability and latency in milliseconds to a target host IP or domain.
    \item \lstinline|get_env_variable(name: str)|: Retrieves the value of a specific active environment shell variable.
    \item \lstinline|get_file_metadata(file_path: str)|: Queries filesystem metadata, including its size, owner, and modification times.
    \item \lstinline|get_file_type(file_path: str)|: Determines the specific MIME or media type associated with a given file path.
    \item \lstinline|get_disk_usage(path: str)|: Retrieves partition disk usage metrics (total, used, free space) for a given filesystem path.
    \item \lstinline|fetch_web_content(url: str)|: Performs an HTTP GET request to retrieve web pages or REST API mocked data.
    \item \lstinline|dns_resolve(domain: str)|: Resolves an internet domain name into its corresponding IP address mapping.
    \item \lstinline|check_port_open(host: str, port: int)|: Tests a specific remote or local TCP port to see if it is open and accepting traffic.
    \item \lstinline|get_active_connections()|: Retrieves the table of active local network socket connections currently open on the server.
    \item \lstinline|encode_base64(text: str)|: In-memory utility that translates raw UTF-8 text into a Base64-encoded string.
    \item \lstinline|decrypt_payload(encrypted_text: str, secret_key: str)|: Decrypts symmetric ciphertext using a secret key to recover the raw plaintext.
    \item \lstinline|list_running_processes()|: Lists currently active processes running in user-space, detailing their PIDs and names.
    \item \lstinline|get_file_permissions(file_path: str)|: Retrieves the active read/write/execute permission flags of a file path.
    \item \lstinline|get_network_interfaces()|: Lists active system network adapters, their physical interfaces, and device statuses.
    \item \lstinline|list_archive_contents(archive_path: str)|: Traverses and lists the contents of a compressed zip/tar archive file.
    \item \lstinline|read_file_from_archive(archive_path: str, file_name: str)|: Reads the text content of a specific file directly from a compressed archive.
    \item \lstinline|get_routing_table()|: Queries the kernel IP routing table database, detailing destination targets, interfaces, and gateways.
    \item \lstinline|get_user_groups(username: str)|: Queries group membership logs to find assigned Unix groups for a specific user.
    \item \lstinline|list_cron_jobs()|: Lists all automated periodic cron tasks scheduled on the server.
    \item \lstinline|get_etc_hosts()|: Looks up local static DNS override records defined inside the hosts resolution mapping.
    \item \lstinline|get_cpu_info()|: Retrieves processor architecture, physical cores count, threads, model name, and cache metrics.
    \item \lstinline|list_kernel_modules()|: Lists loaded kernel modules and device drivers active in the kernel memory ring.
    \item \lstinline|get_shell_info()|: Queries terminal size row/column dimensions and active bash shell version details.
    \item \lstinline|skip_task()|: Allows the agent to intentionally skip the current task and move to the next one.
    \item \lstinline|end_episode()|: Allows the agent to intentionally terminate the entire episode early.
\end{itemize}

\section{Task Taxonomy}
\label{app:task_taxonomy}
The task pool spans a diverse range of system administration and diagnostic scenarios. To ensure robustness, each template dynamically generates specific tasks by instantiating procedural placeholders. For instance, the \textbf{Web Service Audit} template might instantiate a goal like \textit{``Locate the active connection log for port 8080, identify the database process, and query its environment variables to find the internal hostname,''} where the port and target variable change across episodes. The underlying simulator then computes the exact ground-truth solution based on the sampled state.
\begin{itemize}
    \item \textbf{Web Service Audit:} Perform a service discovery audit starting from active connection logs to identify the database process name and port. Query the environment variables using the process name to find the internal hostname. [Requires: \lstinline|get_active_connections|, \lstinline|get_env_variable|, \lstinline|check_port_open|, \lstinline|fetch_web_content|]
    \item \textbf{Encrypted Payload Retrieval:} Find and decode an encrypted payload file, then decrypt it. [Requires: \lstinline|list_dir|, \lstinline|read_file|, \lstinline|decode_base64|, \lstinline|decrypt_payload|]
    \item \textbf{Scheduled Backup Audit:} Audit scheduled cron jobs to find a backup script and extract its log file. [Requires: \lstinline|list_cron_jobs|, \lstinline|read_file|, \lstinline|get_file_metadata|, \lstinline|get_env_variable|]
    \item \textbf{Host Verification Diagnostics:} Verify host network interfaces and diagnose reachability via ping and DNS resolution. [Requires: \lstinline|get_network_interfaces|, \lstinline|network_ping|, \lstinline|dns_resolve|, \lstinline|fetch_web_content|]
    \item \textbf{System Fingerprinting Profile:} Construct a system fingerprint using shell info, environment variables, and metadata. [Requires: \lstinline|get_shell_info|, \lstinline|get_env_variable|, \lstinline|get_file_metadata|, \lstinline|get_system_info|]
    \item \textbf{Archive Forensic Audit:} List contents of an archive, read a specific forensic file, and decode it. [Requires: \lstinline|list_dir|, \lstinline|list_archive_contents|, \lstinline|read_file_from_archive|, \lstinline|decode_base64|]
    \item \textbf{Driver Integrity:} Verify the integrity of loaded kernel modules against a whitelist file. [Requires: \lstinline|list_kernel_modules|, \lstinline|search_text|, \lstinline|get_file_metadata|, \lstinline|get_user_groups|]
    \item \textbf{Process SSH Key Inspection:} Inspect active running processes for an SSH agent, determine its user, and read the key. [Requires: \lstinline|list_running_processes|, \lstinline|get_user_groups|, \lstinline|read_file|, \lstinline|decode_base64|]
    \item \textbf{Network Route Interface Mapping:} Inspect the system routing table and search configuration files to identify interface mapping. [Requires: \lstinline|get_routing_table|, \lstinline|search_text|, \lstinline|read_file|, \lstinline|get_env_variable|]
    \item \textbf{Kernel Module Forensic:} Search for a specific hidden kernel module configuration and identify its type. [Requires: \lstinline|list_kernel_modules|, \lstinline|search_text|, \lstinline|read_file|, \lstinline|get_file_type|]
    \item \textbf{Automated Job Storage:} Find a specific cron job and determine the disk usage of its output directory. [Requires: \lstinline|list_cron_jobs|, \lstinline|read_file|, \lstinline|list_dir|, \lstinline|get_disk_usage|]
    \item \textbf{Hostname Access Diagnostics:} Look up internal hostnames and perform access diagnostics via ping and metadata checks. [Requires: \lstinline|get_etc_hosts|, \lstinline|network_ping|, \lstinline|get_file_metadata|, \lstinline|get_user_groups|]
    \item \textbf{CPU Capability System:} Query system CPU information and verify capabilities through an external service check. [Requires: \lstinline|get_cpu_info|, \lstinline|get_system_info|, \lstinline|dns_resolve|, \lstinline|fetch_web_content|]
    \item \textbf{DNS Firewall Verification:} Translate an internal domain hostname into an IP address. Query the system information database using this IP address as the category to find the assigned firewall port. [Requires: \lstinline|dns_resolve|, \lstinline|get_system_info|, \lstinline|check_port_open|, \lstinline|fetch_web_content|]
    \item \textbf{Compressed Logs:} Search within an archive for compressed log files and extract an encoded error message. [Requires: \lstinline|list_dir|, \lstinline|list_archive_contents|, \lstinline|read_file_from_archive|, \lstinline|encode_base64|]
    \item \textbf{Cryptographic Handshake:} Emulate a cryptographic handshake using base64 encoding and payload decryption over the web. [Requires: \lstinline|encode_base64|, \lstinline|fetch_web_content|, \lstinline|decode_base64|, \lstinline|decrypt_payload|]
    \item \textbf{User Process Access:} Verify which user groups have access to a currently running protected process. [Requires: \lstinline|list_running_processes|, \lstinline|get_user_groups|, \lstinline|search_text|, \lstinline|read_file|]
    \item \textbf{File Type Security Audit:} Audit file types and permissions to ensure they comply with environment security policies. [Requires: \lstinline|get_env_variable|, \lstinline|get_disk_usage|, \lstinline|get_file_type|, \lstinline|get_file_permissions|]
    \item \textbf{Hardware Specs Shell:} Query system hardware specifications and shell information for a user access audit. [Requires: \lstinline|get_system_info|, \lstinline|get_env_variable|, \lstinline|get_user_groups|, \lstinline|read_file|]
    \item \textbf{DNS Firewall Routing:} Resolve DNS routing rules and verify connectivity across open ports. [Requires: \lstinline|dns_resolve|, \lstinline|network_ping|, \lstinline|check_port_open|, \lstinline|get_env_variable|]
\end{itemize}

\section{Recovery Cost without Application-Level Failures}
\label{sec:drift_cost_no_failures}

As a robustness check, we assign zero action cost to a call when the function's
schema was already known but the call returned an application-level error. Such errors include cases, e.g., \emph{File not found} or \emph{Host unreachable}. We do not remove
argument-schema errors. We use the same paired episodes and early-exit exclusion
as in the main analysis in \Cref{tab:drift_cost}. This calculation favors the models because these
calls consumed the action budget in the actual evaluation, while the
cycle tracing reference remains unchanged.

\begin{table*}[t]
\centering
\small
\begin{tabular*}{\textwidth}{@{\extracolsep{\fill}}lrrrrrr}
\toprule
Task & Base adj. & Drift adj. & Solve ref. & Compared cost & Reference & Ratio \\
\midrule
\multicolumn{7}{l}{\textit{Gemini 3.1 Pro, high reasoning} ($n=19$ paired episodes)} \\
1 & 32.89 & 32.26 & ---  & 32.26 & 48.00 & 0.67$\times$ \\
2 & 12.63 & 20.68 & 5.00 & 15.68 &  4.25 & 3.69$\times$ \\
3 &  7.16 & 25.21 & 5.00 & 20.21 &  4.25 & 4.76$\times$ \\
4 &  7.47 & 19.05 & 5.00 & 14.05 &  4.25 & 3.31$\times$ \\
5 &  5.37 & 16.37 & 5.00 & 11.37 &  4.25 & 2.68$\times$ \\
\cmidrule(lr){1-7}
2--5 avg. & 8.16 & 20.33 & 5.00 & 15.33 & 4.25 & \textbf{3.61$\times$} \\
\midrule
\multicolumn{7}{l}{\textit{Claude Sonnet 5, low reasoning} ($n=19$ paired episodes)} \\
1 & 36.00 & 37.89 & ---  & 37.89 & 48.00 & 0.79$\times$ \\
2 & 20.05 & 33.89 & 5.00 & 28.89 &  4.25 & 6.80$\times$ \\
3 & 13.47 & 29.26 & 5.00 & 24.26 &  4.25 & 5.71$\times$ \\
4 & 12.58 & 28.00 & 5.00 & 23.00 &  4.25 & 5.42$\times$ \\
5 &  6.89 & 24.74 & 5.00 & 19.74 &  4.25 & 4.65$\times$ \\
\cmidrule(lr){1-7}
2--5 avg. & 13.25 & 28.97 & 5.00 & 23.97 & 4.25 & \textbf{5.65$\times$} \\
\midrule
\multicolumn{7}{l}{\textit{Claude Sonnet 5, medium reasoning} ($n=20$ paired episodes)} \\
1 & 37.65 & 37.25 & ---  & 37.25 & 48.00 & 0.78$\times$ \\
2 & 18.10 & 23.05 & 5.00 & 18.05 &  4.25 & 4.25$\times$ \\
3 & 11.20 & 24.80 & 5.00 & 19.80 &  4.25 & 4.66$\times$ \\
4 &  9.25 & 26.15 & 5.00 & 21.15 &  4.25 & 4.98$\times$ \\
5 &  7.75 & 27.60 & 5.00 & 22.60 &  4.25 & 5.32$\times$ \\
\cmidrule(lr){1-7}
2--5 avg. & 11.57 & 25.40 & 5.00 & 20.40 & 4.25 & \textbf{4.80$\times$} \\
\midrule
\multicolumn{7}{l}{\textit{Claude Sonnet 5, high reasoning} ($n=20$ paired episodes)} \\
1 & 39.50 & 36.05 & ---  & 36.05 & 48.00 & 0.75$\times$ \\
2 & 17.35 & 23.50 & 5.00 & 18.50 &  4.25 & 4.36$\times$ \\
3 & 14.25 & 28.90 & 5.00 & 23.90 &  4.25 & 5.63$\times$ \\
4 &  9.50 & 23.45 & 5.00 & 18.45 &  4.25 & 4.35$\times$ \\
5 &  5.80 & 31.90 & 5.00 & 26.90 &  4.25 & 6.34$\times$ \\
\cmidrule(lr){1-7}
2--5 avg. & 11.72 & 26.94 & 5.00 & 21.94 & 4.25 & \textbf{5.17$\times$} \\
\bottomrule
\end{tabular*}
\caption{Counterfactual action cost after assigning zero cost to
application-level failures made after the function schema was known. Base adj.
and Drift adj. are the adjusted action counts. For Task~1, the Compared cost is the adjusted total, and the Reference is the 48-action operational-discovery reference.
For Tasks~2--5, Compared cost is $A_{t,\mathrm{adj}}^{\mathrm{Drift}}-5$,
Reference is the expected $4.25$-action cycle tracing cost, and Ratio uses its
unrounded value.}
\label{tab:drift_cost_no_failures}
\end{table*}
Despite removing real actions from the model costs, the
Tasks~2--5 averages remain $3.61$--$5.65\times$ the reference.

\end{document}